\documentclass[10pt,twocolumn,letterpaper]{article}

\usepackage[pagenumbers]{wacv} 

\usepackage{graphicx}
\usepackage{amsmath}
\usepackage{amssymb}
\usepackage{booktabs}
\usepackage{xcolor}
\usepackage[accsupp]{axessibility}  
\usepackage[misc]{ifsym}

\newcommand*{\affaddr}[1]{#1} 

\newcommand*{\email}[1]{\texttt{#1}}

\usepackage[pagebackref,breaklinks,colorlinks]{hyperref}

\usepackage[capitalize]{cleveref}
\crefname{section}{Sec.}{Secs.}
\Crefname{section}{Section}{Sections}
\Crefname{table}{Table}{Tables}
\crefname{table}{Tab.}{Tabs.}

\def\wacvPaperID{137} 
\def\confName{WACV}
\def\confYear{2025}

\begin{document}

\title{360PanT: Training-Free Text-Driven 360-Degree Panorama-to-Panorama Translation}


\author{%
Hai~Wang\thanks{Corresponding author.} \, and Jing-Hao Xue\\
\affaddr{University College London}\\
\email{\tt\small \{hai.wang.22, jinghao.xue\}@ucl.ac.uk}%
}

\maketitle

\begin{abstract}
    Preserving boundary continuity in the translation of 360-degree panoramas remains a significant challenge for existing text-driven image-to-image translation methods. These methods often produce visually jarring discontinuities at the translated panorama's boundaries, disrupting the immersive experience. To address this issue, we propose 360PanT, a training-free approach to text-based 360-degree panorama-to-panorama translation with boundary continuity. Our 360PanT achieves seamless translations through two key components: boundary continuity encoding and seamless tiling translation with spatial control. Firstly, the boundary continuity encoding embeds critical boundary continuity information of the input 360-degree panorama into the noisy latent representation by constructing an extended input image. Secondly, leveraging this embedded noisy latent representation and guided by a target prompt, the seamless tiling translation with spatial control enables the generation of a translated image with identical left and right halves while adhering to the extended input's structure and semantic layout. This process ensures a final translated 360-degree panorama with seamless boundary continuity. Experimental results on both real-world and synthesized datasets demonstrate the effectiveness of our 360PanT in translating 360-degree panoramas. Code is available at \href{https://github.com/littlewhitesea/360PanT}{https://github.com/littlewhitesea/360PanT}.
\end{abstract}

\section{Introduction}
\label{sec:intro}

Text-driven image-to-image (I2I) translation seeks to generate a new image that reflects a given target prompt while following the structure and semantic layout of an input image. For text-driven I2I translation, recent training-free methods, such as Prompt-to-Prompt (P2P) \cite{P2P}, Plug-and-Play (PnP) \cite{PnP} and FreeControl \cite{freecontrol}, are based on pre-trained latent diffusion models (LDMs) \cite{latentdiffusion} and typically employ DDIM inversion \cite{ddim} to obtain the corresponding noisy latent representation of the input image. Subsequently, they leverage attention control \cite{P2P,PnP} or spatial control \cite{freecontrol} to guide the translation process during denoising. By harnessing the powerful generative capabilities of pre-trained LDMs \cite{latentdiffusion}, these methods demonstrate commendable performance in translating ordinary images.

\begin{figure*}[t]
\centering
\vspace{-5mm}
\includegraphics[width=\textwidth]{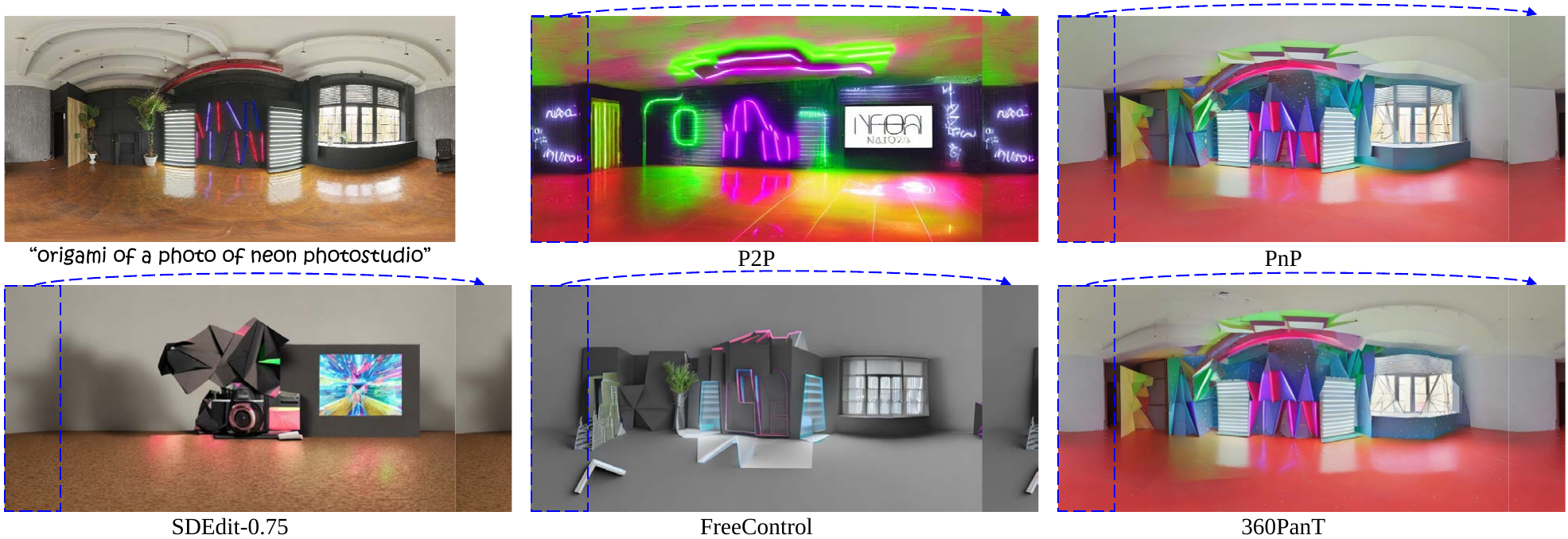}
\caption{\textbf{Example of text-driven 360-degree panorama-to-panorama translation.} To easily identify visual continuity or discontinuity at the boundaries of the translated panoramic image, we copy the left area indicated by the \textcolor{blue}{blue dashed box} and paste it onto the rightmost side of the image. Compared with other methods, our 360PanT performs best in maintaining boundary continuity and preserving the structure and semantic layout of the input 360-degree panorama in the translated result.}
\vspace{-2mm}
\label{fig:teaser}
\end{figure*}

However, directly applying these techniques to 360-degree panoramic images, which are commonly represented by using equirectangular projection \cite{xu2020state}, presents a unique and significant challenge. Unlike ordinary images, 360-degree panoramas possess inherent boundary continuity, where the leftmost and rightmost edges seamlessly connect. Existing I2I translation methods based on DDIM inversion fail to preserve this crucial characteristic, resulting in noticeable discontinuities at the boundaries of translated panoramas, as shown in Figure \ref{fig:teaser}. To solve this problem, we propose 360PanT, a training-free method tailored for text-driven 360-degree panorama-to-panorama (Pan2Pan) translation. Our approach comprises two primary components: \textbf{boundary continuity encoding} and \textbf{seamless tiling translation with spatial control}.

\textbf{Boundary continuity encoding} aims to embed the boundary continuity information of the input 360-degree panorama into the noisy latent representation. This is achieved by first creating an extended input image obtained from splicing two copies of the original input panorama. This extended input is then processed by the encoder of a pre-trained LDM. Finally, DDIM inversion is applied to the resulting latent feature, yielding a noisy latent feature that intrinsically encodes the boundary continuity.

While one might consider directly applying existing state-of-the-art (SOTA) I2I translation techniques like PnP~\cite{PnP} or FreeControl \cite{freecontrol} to this noisy latent feature, such an approach presents two significant drawbacks. Firstly, processing the entire noisy latent feature on a single high-end GPU (e.g., 24GB) leads to out-of-memory errors. Secondly, these SOTA methods cannot guarantee the preservation of identical left and right halves throughout the denoising process, potentially disrupting the 360-degree panoramic structure.

To address these issues, we put forward \textbf{seamless tiling translation with spatial control}. Specifically, we leverage a key property of StitchDiffusion \cite{stitchdiffusion}, a method designed for generating 360-degree panoramas from using a customized latent diffusion model \cite{latentdiffusion}. StitchDiffusion inherently produces images with identical left and right halves, ensuring panoramic continuity. Moreover, cropped patches of the noisy latent feature, instead of the entire noisy latent feature, are independently processed within StitchDiffusion during denoising. This \textbf{seamless tiling translation} strategy effectively addresses the memory constraints and guarantees the preservation of the 360-degree panoramic structure.

However, relying solely on the noisy latent feature and the target prompt leads to a translated image that deviates from the structure and semantic layout of the extended input. To solve this problem, we integrate \textbf{spatial control} into the seamless tiling translation process. Inspired by PnP \cite{PnP}, we inject spatial features and self-attention maps from the extended input image into the seamless tiling translation process. The spatial control mechanism enables the translated image to maintain the structure and semantic layout of the extended input, resulting in a finely translated 360-degree panorama.

Furthermore, an alternative to spatial feature and self-attention map injection is explored. Drawing inspiration from FreeControl \cite{freecontrol}, we introduce structure guidance and appearance guidance into the seamless tiling translation process. This approach allows our 360PanT to support a variety of 360-degree panoramic maps (e.g., segmentation masks and edge maps) as input conditions instead of a standard 360-degree panoramic image.

\noindent
\textbf{Novelties and Contributions.} (1) We propose 360PanT, the first training-free method for text-driven 360-degree panorama-to-panorama translation, which consists of two key components: boundary continuity encoding and seamless tiling translation with spatial control. (2) Beyond standard 360-degree panoramic images, 360PanT can expand its capacity to support various types of 360-degree panoramic maps (e.g., segmentation masks and edge maps) as input conditions. This flexibility extends its applications to various scenarios requiring diverse input formats. (3) Extensive experiments on both real-world and synthesized datasets demonstrate the effectiveness of our proposed method in translating 360-degree panoramas through text prompts.


\section{Related Work}

\noindent
\textbf{Text-Driven 360-Degree Panorama Generation.} 
The objective of text-driven panorama generation \cite{diffcollage,multidiffusion,syncdiffusion,panogen} is to produce panoramic images aligned with given textual descriptions. Unlike ordinary panoramic images, 360-degree panoramic images offer immersive experiences and find broad applications in virtual reality \cite{virtual}, autonomous driving \cite{autonomous}, and indoor design \cite{indoor}.

For synthesizing 360-degree panoramas from text prompts, Text2Light \cite{text2light} introduces a hierarchical framework comprising a dual-codebook discrete representation, a text-conditioned global sampler, and a structure-aware local sampler. In contrast, recent approaches \cite{stitchdiffusion,diffusion360,panfusion,lu2024autoregressive,MVDiffusion} explore text-to-image latent diffusion models \cite{latentdiffusion} for text-driven 360-degree panorama generation. Among these methods, StitchDiffusion \cite{stitchdiffusion} proposes additional denoising twice on the {stitch patch} based on MultiDiffusion \cite{multidiffusion}, ensuring the generated image to have identical left and right halves. We leverage this crucial attribute of StitchDiffusion to achieve seamless tiling translation in our 360PanT.

\begin{figure*}[t]
\centering
\vspace{-3mm}
\includegraphics[width=\textwidth]{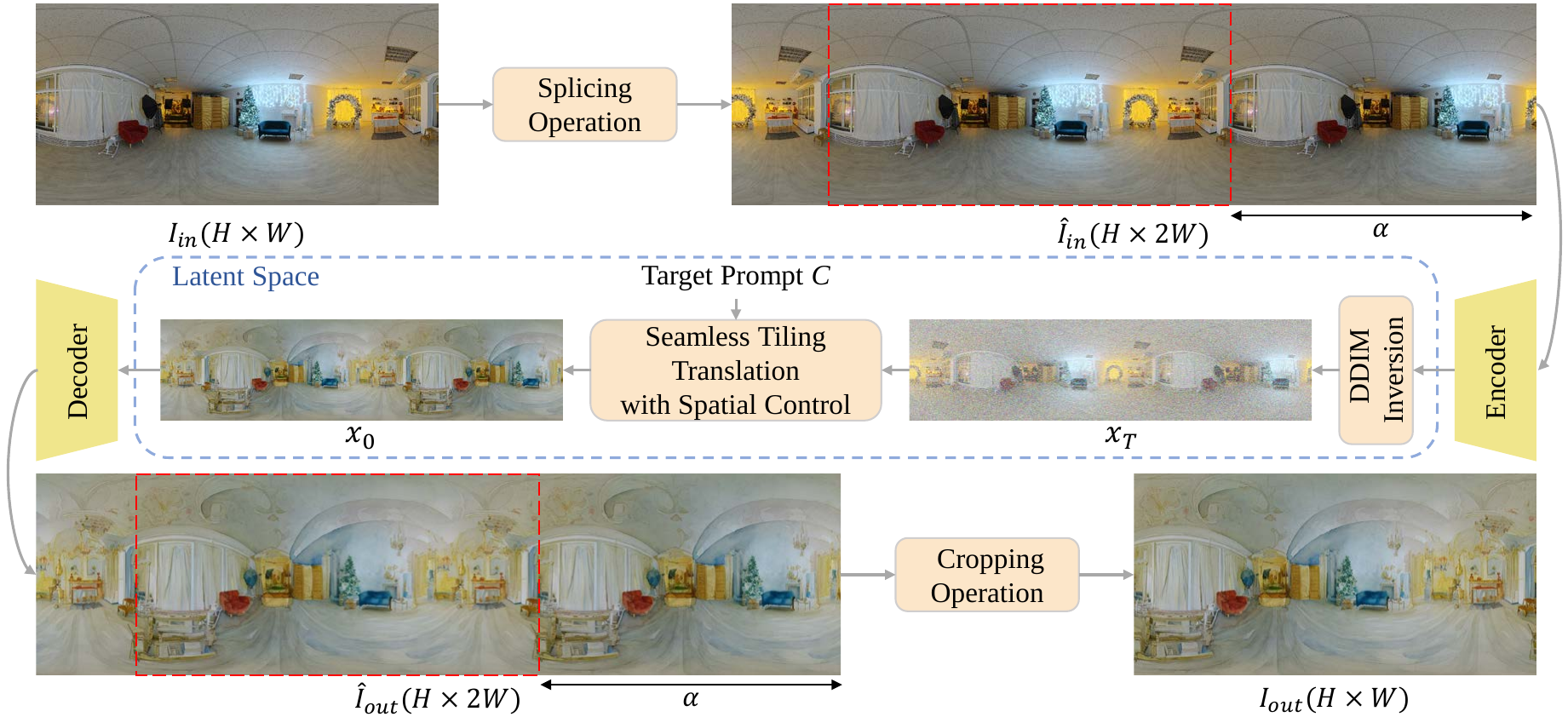}
\caption{\textbf{Method overview.} Our 360PanT comprises two primary components: boundary continuity encoding and seamless tiling translation with spatial control. The boundary continuity encoding component embeds the boundary continuity information of $I_{in}$ into the noisy latent feature $x_{T}$. Subsequently, guided by the target prompt $C$, $x_T$ undergoes seamless tiling translation with spatial control to produce the denoised translated latent feature $x_0$. Finally, the translated 360-degree panorama $I_{out}$, aligned with the target prompt $C$, is achieved by cropping from the translated image $\hat{I}_{out}$.}
\vspace{-1mm}
\label{fig:framework}
\end{figure*}

\noindent
\textbf{Text-Guided Image-to-Image Translation.} Image-to-image (I2I) translation aims to learn a mapping that transforms images between domains while maintaining the semantic layout and structure of the input image. Over the past few years, GAN-based I2I translation methods have been extensively investigated \cite{pix2pix,CycleGAN2017,DiscoGAN,unit,stargan,munit,tsit,dwc-gan,psp,vqgan-clip}. Recently, diffusion models \cite{sohl2015deep,ddpm,song2020score,ddim,dhariwal2021diffusion} have emerged as a powerful alternative to GANs \cite{goodfellow2020generative}, exhibiting superior performance in image synthesis. This shift has motivated research into exploring diffusion models for I2I translation \cite{palette,wang2022pretraining,diffuseit,diffusionclip,PnP,P2P,freecontrol,sdedit,bbdm,zeroi2i}. 

Notably, training-free text-driven I2I translation methods \cite{sdedit,zeroi2i,P2P,PnP,freecontrol}, building upon pre-trained latent diffusion models (LDMs) \cite{latentdiffusion}, have gained significant attention. For example, Plug-and-Play (PnP) \cite{PnP} proposes to inject spatial features and self-attention maps into the denoising process of the translated image for enhancing structure preservation. Different from PnP, FreeControl \cite{freecontrol} introduces appearance guidance and structure guidance to achieve spatial control of the translated image. Leveraging the powerful generative capabilities of pre-trained LDMs, these text-driven I2I methods achieve impressive results on ordinary images. However, when applied to 360-degree panoramic images, they fail to maintain visual continuity at the boundaries of the translated images. To address this problem, we propose a training-free method called 360PanT. By using our designed boundary continuity encoding and seamless tiling translation with spatial control, 360PanT successfully achieves the translation of 360-degree panoramas.


\section{Methodology}

The framework of our 360PanT is illustrated in Figure~\ref{fig:framework}, consisting of two key components: boundary continuity encoding and seamless tiling translation with spatial control. Details of each component are elaborated in the following.

\subsection{Boundary Continuity Encoding}

Recent training-free text-driven image-to-image (I2I) translation methods, such as Prompt-to-Prompt (P2P) \cite{P2P}, Plug-and-Play (PnP) \cite{PnP} and FreeControl \cite{freecontrol}, face inherent limitations when applied to 360-degree panoramas. This limitation stems from the inability of the DDIM inversion process \cite{ddim}, a core component of these methods, to encode the continuous information between the leftmost and rightmost sides of a 360-degree panorama. DDIM inversion, primarily designed for ordinary images, converts a clear image into a noisy latent representation without accounting for the cyclical nature of 360-degree panoramas. Consequently, these training-free I2I translation methods \cite{sdedit,P2P,PnP,freecontrol} relying on DDIM inversion fail to maintain visual continuity between the edges of the final translated panorama.

To address this challenge, we propose a straightforward yet effective method to encode this crucial continuous information. Our approach involves firstly splicing two identical copies of the input panorama, to create an extended image that serves as input for the DDIM inversion process \cite{ddim}. Formally, given an input 360-degree panorama $I_{in}$ with dimensions $3\times{H}\times{W}$, the extended input $\hat{I}_{in}$ with dimensions $3\times{H}\times{2W}$ is constructed as follows:
\begin{equation}\label{eq:extended}
\hat{I}_{in} = Splice(I_{in}[:,:,\alpha: W], I_{in}, I_{in}[:,:, 0:\alpha]) \ ,
\end{equation}
where $\alpha$ is a split constant controlling the splicing point, and \emph{Splice} denotes the image splicing operation. Note that (1) setting $\alpha$ equal to $W$ results in $\hat{I}_{in}$ being a direct concatenation of two copies of $I_{in}$; and (2) the extended input $\hat{I}_{in}$ consistently maintains identical left and right halves regardless of the value of $\alpha$. Subsequently, this extended image $\hat{I}_{in}$ is encoded into the latent space, and DDIM inversion is applied to its corresponding latent feature, yielding a noisy latent feature $x_{T}$ with dimensions $4\times\frac{H}{8}\times\frac{2W}{8}$, which naturally embeds the boundary continuous information of the original 360-degree panorama.

\subsection{Seamless Tiling Translation}

At this stage, we have a noisy latent feature $x_{T}$ including the continuous information of the original 360-degree panorama $I_{in}$. A direct approach to performing training-free text-driven panorama-to-panorama translation would be to apply existing I2I translation methods, such as PnP~\cite{PnP} or FreeControl \cite{freecontrol}, to $x_{T}$ and then crop the translated image (with dimensions $3\times{H}\times{2W}$) to obtain the final 360-degree output. However, this approach has two significant drawbacks: (1) directly processing the entire $x_{T}$ on a single high-end GPU (e.g., 24GB) results in out-of-memory errors; and (2) these methods cannot ensure that the translated image will still maintain identical left and right halves during the denoising process, potentially disrupting the panoramic structure.

To overcome these issues, we leverage a key property of StitchDiffusion \cite{stitchdiffusion}, a method designed for generating 360-degree panoramas using a customized latent diffusion model \cite{latentdiffusion}. StitchDiffusion inherently produces images with identical left and right halves by design, ensuring the preservation of the 360-degree panoramic structure. Furthermore, at denoising step $t$, where $t\in\{T, T-1, \cdots, 1\}$, cropped patches of $x_t$, rather than the entire $x_t$, are independently processed within StitchDiffusion. Therefore, instead of directly applying existing I2I translation methods, we employ StitchDiffusion to translate the noisy latent feature $x_{T}$. This approach effectively addresses the aforementioned memory constraints and ensures the translated image maintaining identical left and right halves.

Specifically, at denoising step $t$, the noisy latent feature $x_t$ is divided into $n$ overlapping patches. Let $F_{i}(x_{t})$ represent the $i$-th cropped patch of size $\frac{H}{8}\times\frac{W}{8}$, where $i\in\left\{1, 2, \cdots, n\right\}$. Here, the mapping $F_{i}$ denotes the cropping operation for the $i$-th patch, and its inverse mapping, $F_i^{-1}$, places the patch back into its original position. The number of patches, $n$, is determined by $\frac{W}{8\omega} + 1$, where $\omega$ indicates the sliding distance between adjacent patches $F_i(x_t)$ and $F_{i+1}(x_t)$. In addition, let $\Phi$ and $C$ denote a pre-trained latent diffusion model  \cite{latentdiffusion} and a target prompt, respectively. In this situation, the sequential denoising process of a training-free I2I translation using StitchDiffusion, termed seamless tiling translation process, can be represented as
\begin{equation}\label{eq:stitch}
\small
\begin{split}
x_{t-1} &= \sum_{j=1}^{2}\frac{{^{j}F^{-1}_{n+1}}(\mathbf{1})}{\Pi} \otimes {^{j}F^{-1}_{n+1}}(\Phi({^{j}F_{n+1}}(x_t),C)) \\ 
&+ \sum_{i=1}^{n}\frac{F^{-1}_i(\mathbf{1})}{\Pi} \otimes F_i^{-1}(\Phi(F_i(x_t),C))
  \ ,
\end{split}
\end{equation}
where ${^{j}F_{n+1}}(\cdot)$ and ${^{j}F}^{-1}_{n+1}(\cdot)$ are the $j$-th additional mapping and inverse mapping of the stitch patch, respectively; and $\Pi$ denotes ${{^{1}{F}}_{n+1}^{-1}(\mathbf{1}) + {^{2}{F}}_{n+1}^{-1}(\mathbf{1}) + \sum_{i=1}^{n}F_{i}^{-1}(\mathbf{1})}$, where $\mathbf{1}$ refers to a latent feature with dimensions $4\times\frac{H}{8}\times\frac{W}{8}$ with all values equal to 1. {The stitch patch, a special cropped patch,} is defined as $Splice(x_t[:, :, \frac{3W}{16}:\frac{2W}{8}], x_t[:,:,0:\frac{W}{16}])$, where, as in Eq.~\ref{eq:extended}, $Splice$ is the splicing operation.

Through the seamless tiling translation process (Eq.~\ref{eq:stitch}), we obtain the final denoised latent feature $x_0$ with dimensions $4\times\frac{H}{8}\times\frac{2W}{8}$. Consequently, the corresponding translated image $\hat{I}_{out}$ with dimensions $3\times{H}\times{2W}$ decoded from $x_0$ maintains identical left and right halves while corresponding to the target prompt $C$.

\subsection{Seamless Tiling Translation with Spatial Control}

Capitalizing on both boundary continuity encoding and seamless tiling translation, the diffusion model $\Phi$ can produce a translated image $\hat{I}_{out}$ with identical left and right halves, aligned with the target prompt $C$. However, the seamless tiling translation relies solely on $C$ and the initial noisy latent feature $x_{T}$. Consequently, the translated image $\hat{I}_{out}$ may not fully adhere to the structure and semantic layout of the extended input $\hat{I}_{in}$. To address this issue, we propose to incorporate spatial control into the seamless tiling translation, enabling training-free text-based 360-degree panorama-to-panorama (Pan2Pan) translation. 

Specifically, following the Plug-and-Play (PnP) method~\cite{PnP}, we inject spatial features $\textbf{\emph{f}}_t$ and self-attention maps $\textbf{\emph{A}}_t$ from $x_{t-1}^o = \Phi(x_{t}^o,\varnothing)$ into the seamless tiling translation process, where $t\in\{T, T-1, \cdots, 1\}$. Here, $x_{T}^o$ is identical to $x_T$, and $\varnothing$ represents a null text prompt. In this context, the seamless tiling translation process with spatial control is given by
\begin{equation}\label{eq:stitch_pnp}
\small
\begin{split}
x_{t-1} &= \sum_{j=1}^{2}\frac{{^{j}F^{-1}_{n+1}}(\mathbf{1})}{\Pi} \otimes {^{j}F^{-1}_{n+1}}(\Phi({^{j}F_{n+1}}(x_t),C; \textbf{\emph{f}}_t, \textbf{\emph{A}}_t)) \\ 
&+ \sum_{i=1}^{n}\frac{F^{-1}_i(\mathbf{1})}{\Pi} \otimes F_i^{-1}(\Phi(F_i(x_t),C; \textbf{\emph{f}}_t, \textbf{\emph{A}}_t))
  \ .
\end{split}
\end{equation}
Utilizing this spatially controlled translation process, we decode the final denoised latent feature $x_0$ to get the translated image $\hat{I}_{out}$ with dimensions $3\times{H}\times{2W}$. Subsequently, we extract the final translated 360-degree panorama $I_{out}$ with dimensions $3\times{H}\times{W}$ by cropping $\hat{I}_{out}$:
\begin{equation}\label{eq:pnp_cropping}
I_{out} = \hat{I}_{out}[:, :, W-\alpha:2W-\alpha] \ ,
\end{equation}
where, as in Eq. \ref{eq:extended}, $\alpha$ is the split constant.

To further enhance 360PanT's versatility and enable support for diverse input conditions (e.g., segmentation masks and edge maps) beyond using standard 360-degree panoramic images, we explore an alternative to spatial feature and self-attention map injection. Inspired by FreeControl \cite{freecontrol}, we introduce structure guidance $g_s(t)$ and appearance guidance $g_a(t)$ into the seamless tiling translation process, where $t\in\{T, T-1, \cdots, 1\}$. These guidance terms, $g_s(t)$ and $g_a(t)$, are derived from the denoising process of $x_t$ and $x^{r}_t$, respectively. Here, $x^{r}_T$ is a randomly initialized latent feature following a normal distribution, which is not equal to $x_T$. In this context, the seamless tiling translation process incorporating FreeControl's spatial control is updated as
\begin{equation}\label{eq:stitch_freecontrol}
\small
\begin{split}
x^{r}_{t-1} &= \sum_{j=1}^{2}\frac{{^{j}F^{-1}_{n+1}}(\mathbf{1})}{\Pi} \otimes {^{j}F^{-1}_{n+1}}(\Phi({^{j}F_{n+1}}(x^{r}_t),C; g_a(t), g_s(t))) \\ 
&+ \sum_{i=1}^{n}\frac{F^{-1}_i(\mathbf{1})}{\Pi} \otimes F_i^{-1}(\Phi(F_i(x^{r}_t),C; g_a(t), g_s(t)))
  \ .
\end{split}
\end{equation}
Note that this seamless tiling translation process is performed on the latent feature $x^{r}_T$ instead of $x_T$ to support diverse input conditions. Similarly, we obtain the final translated image $\hat{I}_{out}$ with dimensions $3\times{H}\times{2W}$ decoded from $x^{r}_{0}$. A cropping operation is then carried out to achieve the corresponding translated 360-degree panorama $I_{out}$, as described in Eq. \ref{eq:pnp_cropping}.


\section{Experiments and Results}

\noindent
\textbf{Implementation details.} The values of $H$ and $W$ in this paper are 512 and 1024. We set the values of split constant $\alpha$ and sliding distance $\omega$ to 768 and 16, respectively. The version of the pre-trained latent diffusion model \cite{latentdiffusion} is Stable Diffusion 2-1-base. For seamless tiling translation with spatial control, our 360PanT method primarily employs PnP's spatial control mechanism \cite{PnP}. To enable support for diverse input conditions, we introduce a variant denoted as 360PanT (F), which utilizes FreeControl's spatial control \cite{freecontrol} instead of PnP. The settings for the spatial control components and denoising steps $T$ within 360PanT and 360PanT (F) are consistent with the default settings of PnP and FreeControl, respectively. All experiments were carried out using a single NVIDIA L4 GPU.

\noindent
\textbf{Datasets.} Our 360PanT is capable of translating both real-world and synthesized 360-degree panoramas guided by text prompts. Due to the absence of a benchmark dataset for text-driven 360-degree panorama-to-panorama (Pan2Pan) translation, we established two datasets for this purpose. The first dataset, termed \emph{360PanoI-Pan2Pan}, is derived from the \emph{360PanoI} dataset \cite{stitchdiffusion}, which contains 120 real-world 360-degree panoramas across eight scenes. Complementing this, we created \emph{360syn-Pan2Pan}, a synthesized dataset comprising 120 360-degree panoramic images generated using the method outlined in \cite{stitchdiffusion}. To construct text-image pairs for 360-degree Pan2Pan translation, we defined 10 translation types (e.g., watercolor painting, anime artwork, and cartoon). The target prompt for each input 360-degree panorama was formed by randomly selecting a translation type and combining it with the original text prompt. For further details on the target prompts for the two datasets, please refer to the supplementary material. 

\noindent
\textbf{Evaluation metrics.} To quantitatively evaluate the effectiveness of various methods for text-driven 360-degree Pan2Pan translation, we employ metrics used in PnP~\cite{PnP}. Specifically, we utilize text and image encoders from CLIP~\cite{clip} to extract textual embeddings of target prompts and image embeddings of corresponding translated panoramic images. We then calculate the average cosine similarity, which is referred to as \emph{CLIP-score}, between these textual and image embeddings. In addition, we use the DINO-ViT self-similarity distance \cite{dino-vit}, denoted as \emph{DINO-score}, to assess the preservation of structural integrity in the translated 360-degree panoramas compared to the input 360-degree panoramas. These two metrics are reported for the \emph{360PanoI-Pan2Pan} and \emph{360syn-Pan2Pan} datasets.

\begin{figure*}[t]
\centering
\vspace{-6mm}
\includegraphics[width=\textwidth]{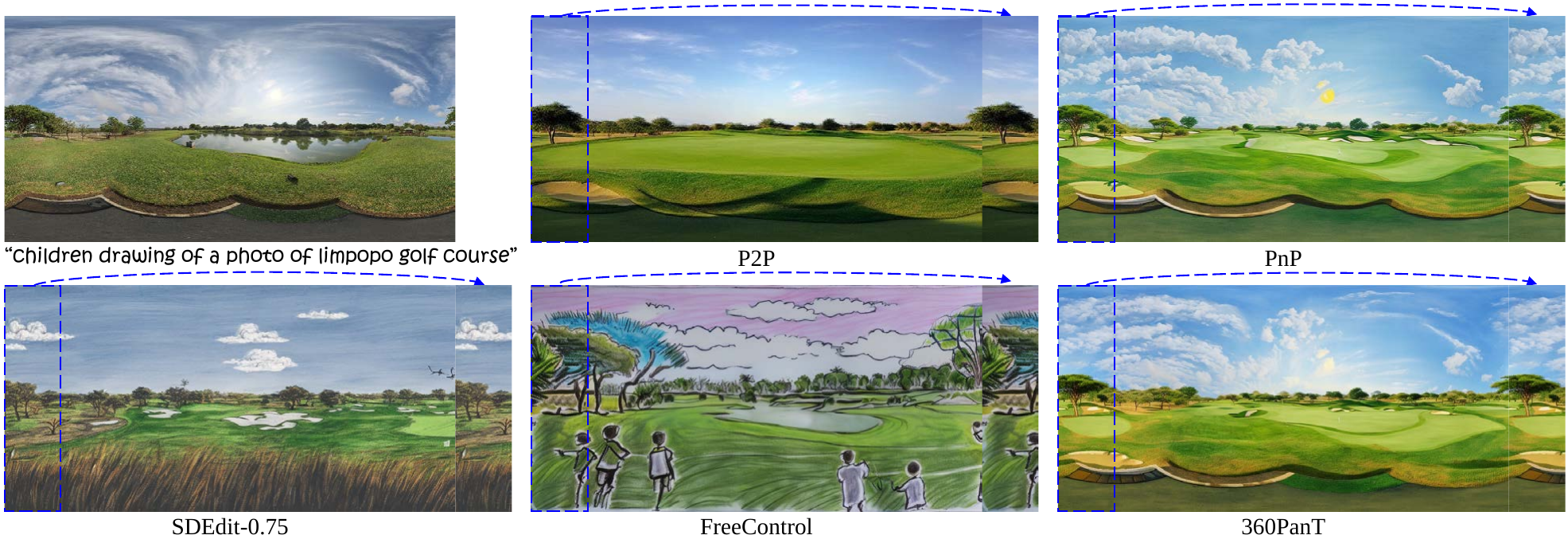}
\caption{\textbf{Visual results on real-world 360-degree panorama.} To easily identify visual continuity or discontinuity at the boundaries, we copy the left area of the panorama indicated by the \textcolor{blue}{blue dashed box} and paste it onto the rightmost side of the image. Current I2I translation methods fail to maintain visual continuity at the boundaries of the translated panoramas. In contrast, our 360PanT not only ensures boundary continuity but also preserves the guidance structure in the translated 360-degree output.}
\vspace{-2mm}
\label{fig:real-world}
\end{figure*}

\begin{figure*}[t]
\centering
\vspace{-2mm}
\includegraphics[width=\textwidth]{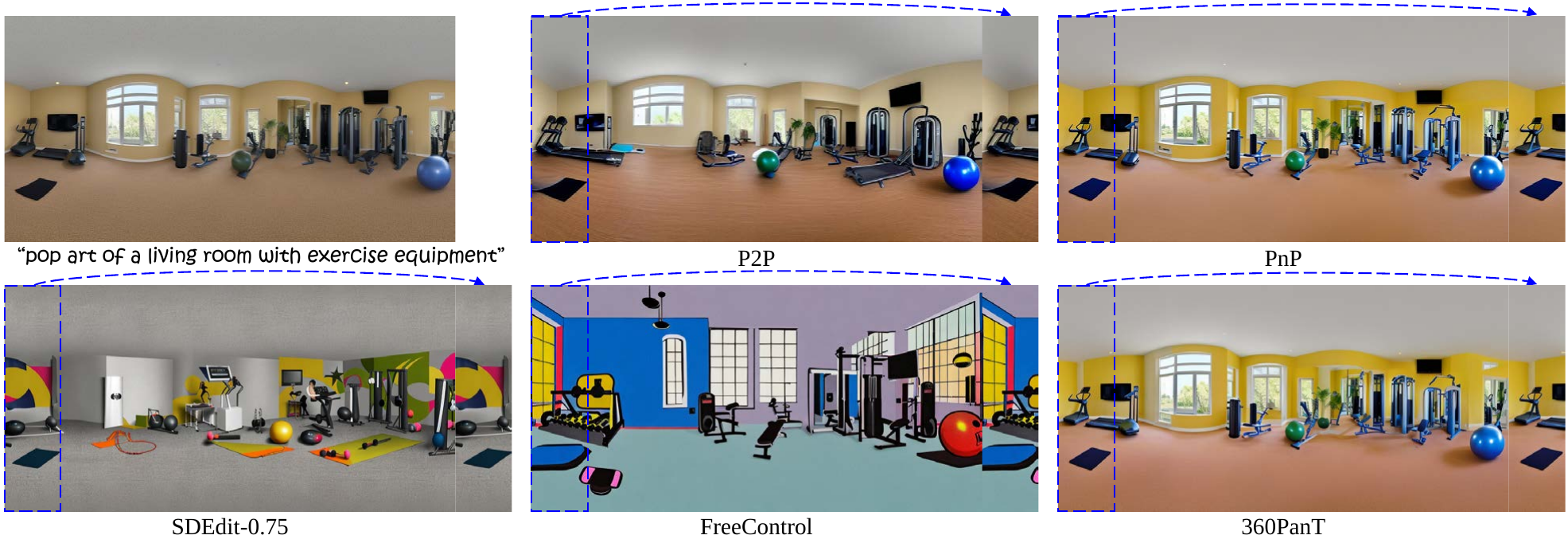}
\caption{\textbf{Visual results on synthesized 360-degree panorama.} Compared to other text-driven I2I methods, our 360PanT performs better in maintaining the visual continuity at the boundaries while also adhering to the structure and semantic layout of the input 360-degree panoramic image. For more visual results, please refer to the supplementary material.}
\vspace{-2mm}
\label{fig:syn}
\end{figure*}

\begin{figure}[t]
\centering
\includegraphics[width=\columnwidth]{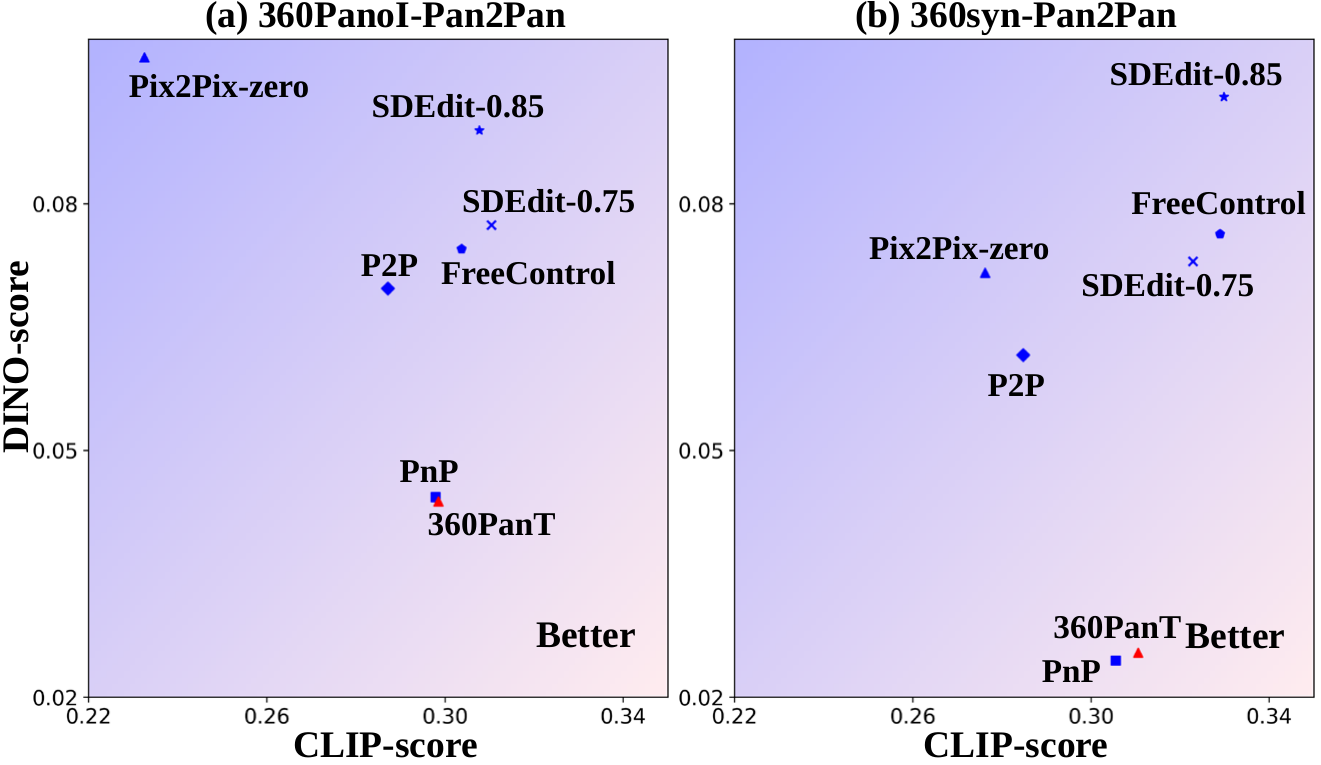}
\caption{\textbf{Quantitative comparison.} DINO-score (lower is better) is to evaluate the structure preservation, while CLIP-
score (higher is better) is to assess the prompt fidelity. Bottom-right is the best.}
\label{fig:clip_dino}
\end{figure}

\subsection{Comparisons with Other Methods}

We compare our 360PanT with state-of-the-art (SOTA) text-driven image-to-image (I2I) translation approaches: SDEdit \cite{sdedit}, Pix2Pix-zero \cite{zeroi2i}, Prompt-to-Prompt (P2P) \cite{P2P}, Plug-and-Play (PnP) \cite{PnP}, FreeControl \cite{freecontrol}. Visual results from the different methods on the translation of real-world and synthesized 360-degree panoramas are illustrated in Figure \ref{fig:real-world} and Figure \ref{fig:syn}, respectively. These figures demonstrate that these SOTA text-driven I2I translation methods fail to preserve the boundary continuity in the translated panoramas. In contrast, our 360PanT not only successfully maintains the visual continuity at the boundaries of the translated panoramas, but also ensures the translated results adhere to the structure and semantic layout of the input 360-degree panoramas. Note that due to space limitations, we only present part visual results here; additional visual results are in the supplementary material.

To further evaluate the performance, we analyze the \emph{CLIP-score} and \emph{DINO-score} metrics across the \emph{360PanoI-Pan2Pan} and \emph{360syn-Pan2Pan} datasets. The results, depicted in Figure \ref{fig:clip_dino}, reveal a close alignment between PnP and 360PanT in both metrics. This similarity is expected, given that 360PanT adopts the same spatial control as PnP. However, a key limitation of PnP is its inability to maintain visual continuity at panorama boundaries. Conversely, our 360PanT can produce translated panoramas with continuous boundaries.

\subsection{Ablation Studies}

\begin{figure}[t]
\centering
\vspace{-8mm}
\includegraphics[width=\columnwidth]{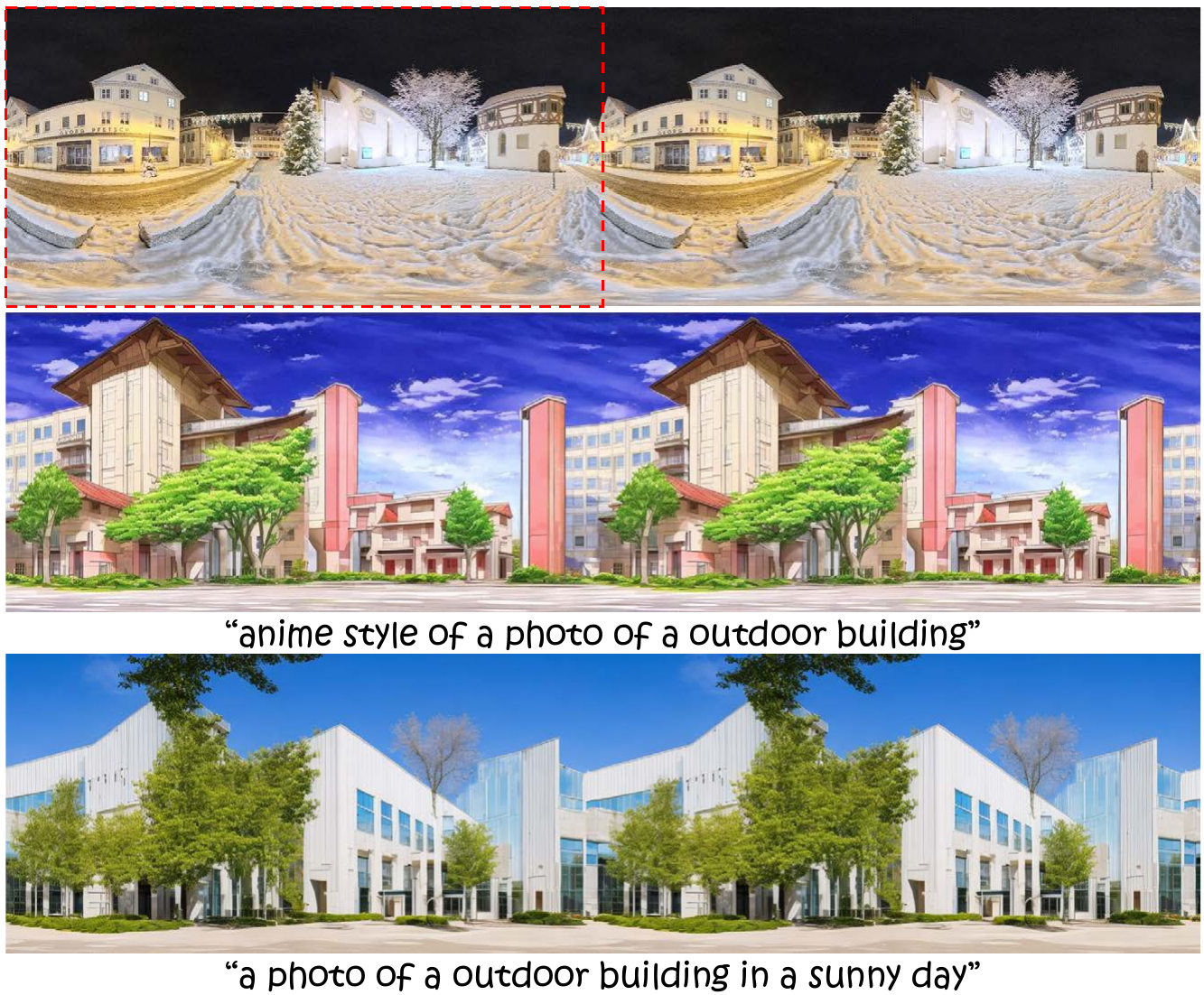}
\caption{\textbf{Ablation on seamless tiling translation effect.} The image in
the first row is the extended input, with the input 360-degree
panorama highlighted within the \textcolor{red}{red dashed box}. The second and third rows display the translated images using seamless tiling
translation with distinct target prompts. Notably, both translated images exhibit identical left and right halves, effectively demonstrating the seamless tiling effect, while simultaneously corresponding to their respective target prompts.}
\vspace{-5mm}
\label{fig:seamless}
\end{figure}

\noindent
\textbf{Effect of seamless tiling translation.} To demonstrate the effectiveness of seamless tiling translation, we conducted some simple I2I translation experiments. Specifically, with a 360-degree panorama denoted as $I_{in}$ with dimensions $3\times512\times1024$ (indicated by the red dashed box in Figure \ref{fig:seamless}), two identical copies were directly spliced to generate an extended image $\hat{I}_{in}$. Subsequently, DDIM inversion \cite{ddim} was applied to the latent feature representation of $\hat{I}_{in}$. The resulting noisy latent feature, $x_T$, underwent seamless tiling translation (Eq. \ref{eq:stitch}) guided by two distinct text prompts, respectively. This process yielded two corresponding translated images. Figure \ref{fig:seamless} illustrates the successful generation of two translated images with dimensions $3\times512\times2048$. These images exhibit identical left and right halves while corresponding to their respective target prompts, highlighting the efficacy of the seamless tiling translation.

\begin{figure}[t]
\centering
\vspace{-8mm}
\includegraphics[width=\columnwidth]{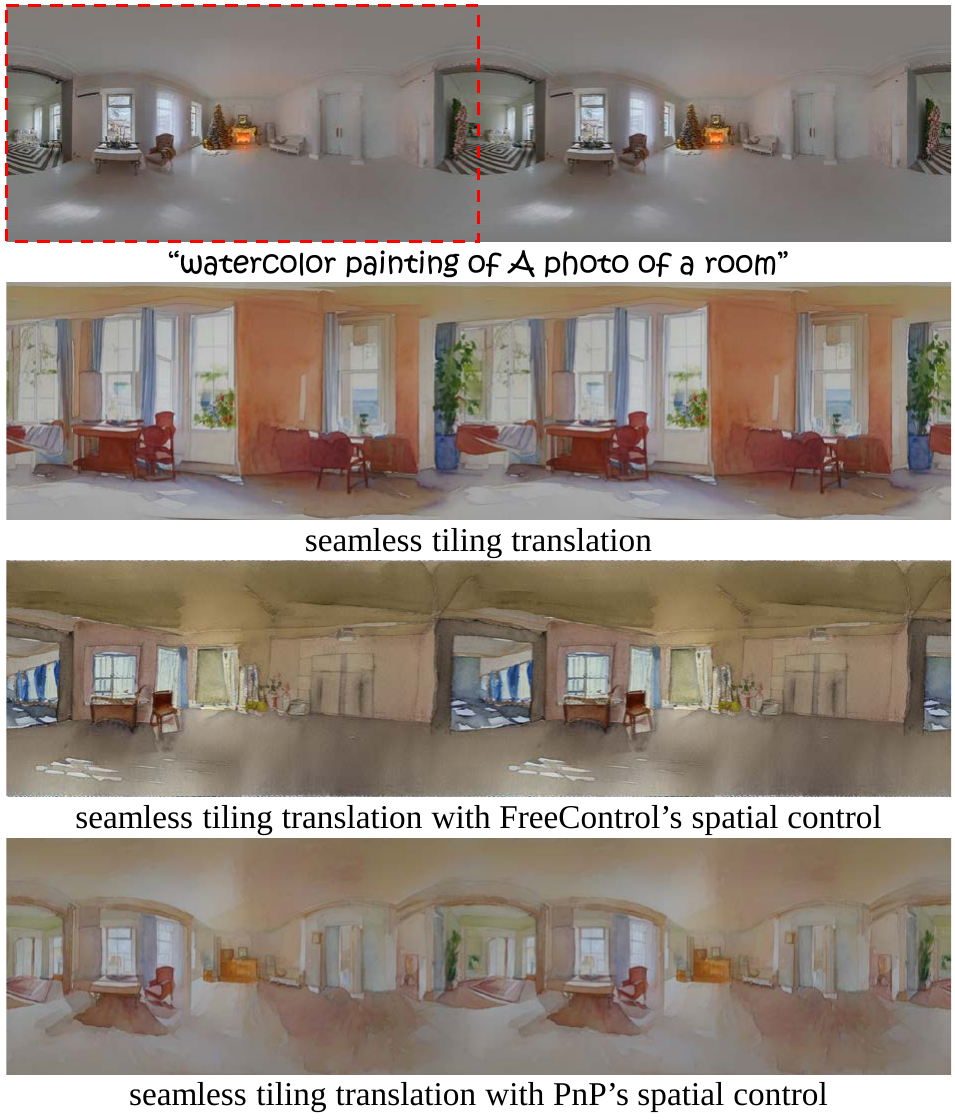}
\caption{\textbf{Ablation on spatial control for seamless tiling translation.} The first row displays the extended input $\hat{I}_{in}$, with the original input 360-degree
panorama highlighted within the \textcolor{red}{red dashed box}. Subsequent rows present the translated images generated by using the same target prompt but employing the following different methods: (2nd row) seamless tiling translation alone; (3rd row) seamless tiling translation with FreeControl's spatial control; and (4th row) seamless tiling translation with PnP's spatial control. Visual comparison discloses that integrating FreeControl's spatial control enhances the preservation of structure and semantic layout from $\hat{I}_{in}$; and incorporating PnP's spatial control improves preservation even more than FreeControl's approach.}
\vspace{-5mm}
\label{fig:seamless_spatial}
\end{figure}

\begin{figure*}[t]
\centering
\vspace{-5mm}
\includegraphics[width=\textwidth]{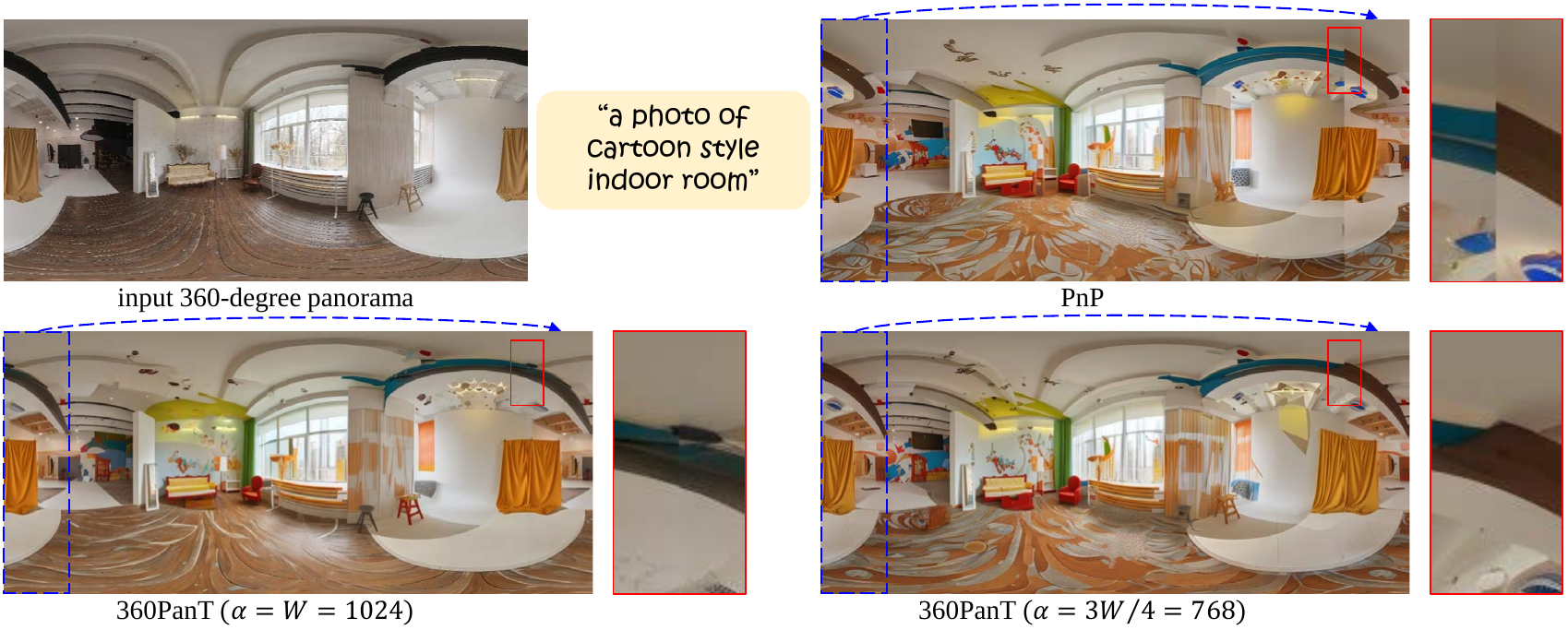}
\caption{\textbf{Ablation on choice of split constant $\alpha$.} While 360PanT with $\alpha=W$ significantly improves the boundary continuity of the translated panorama compared with PnP, a minor crack artifact is still noticeable in the stitched area upon closer inspection (see zoomed-in region highlighted by the \textcolor{red}{red solid box}). In contrast, setting $\alpha$ to $\frac{3W}{4}$ in 360PanT yields a panorama without visible crack artifacts in the stitched region. A further explanation of this parameter choice is available in the supplementary material.}
\vspace{-1mm}
\label{fig:alpha_choice}
\end{figure*} 

\begin{figure*}[t]
\centering
\vspace{-2mm}
\includegraphics[width=\textwidth]{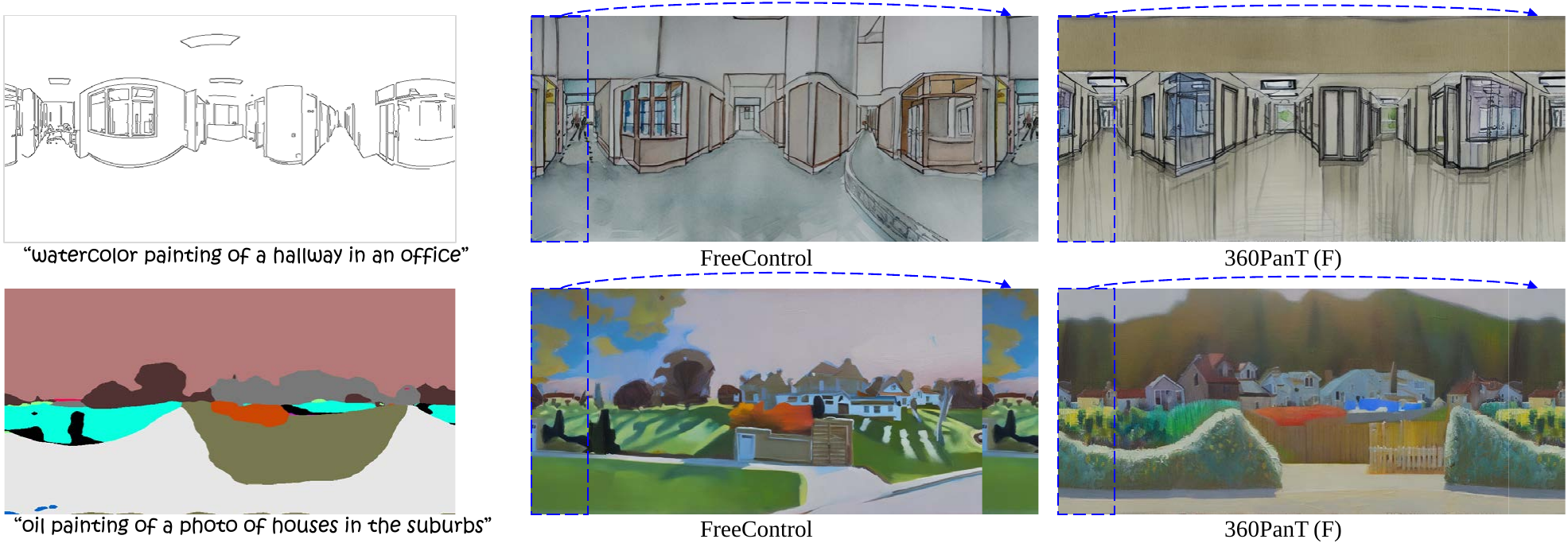}
\caption{\textbf{Visual results using other control conditions.} FreeControl is unable to guarantee the boundary continuity of the translated panoramas. In contrast, our 360PanT (F) enables the translated 360-degree panoramas with continuous boundaries regardless of the input conditions. For more visual results, please refer to the supplementary material.}
\vspace{-3mm}
\label{fig:multiple}
\end{figure*}

\noindent
\textbf{Seamless tiling translation with spatial control.} To investigate the impact of spatial control mechanisms on seamless tiling translation, we carried out a comparative experimental study. In this study, we utilized an extended input image $\hat{I}_{in}$ for translation guided by a target prompt. This image underwent three distinct translation processes: (1) seamless tiling translation alone, (2) seamless tiling translation with PnP's spatial control, and (3) seamless tiling translation with FreeControl's spatial control. The resulting translated images are displayed in Figure \ref{fig:seamless_spatial}. 

We observe that, firstly, incorporating FreeControl's spatial control into seamless tiling translation improves the translated image's adherence to the structure and semantic layout of the extended input image $\hat{I}_{in}$, compared with seamless tiling translation alone. Secondly, integrating PnP's spatial control into seamless tiling translation preserves the structure and semantic layout of $\hat{I}_{in}$ even more effectively than using FreeControl's spatial control. Based on these findings, we adopt PnP's spatial control in our 360PanT method. To distinguish between these variations, we refer to 360PanT incorporating FreeControl's spatial control as 360PanT (F) throughout this paper. While 360PanT (F) is not so effective as 360PanT in structure preservation, it enables support for various input conditions beyond using standard 360-degree panoramic images, as described in Section~\ref{subsection4.3}.


\noindent
\textbf{Choice of split constant $\alpha$.} To study the influence of parameter $\alpha$ on the quality of the final translated 360-degree panorama, we carried out experiments on our 360PanT with varying $\alpha$ values: $W$ (1024) and $\frac{3W}{4}$ (768), where $W$ denotes the width of the input 360-degree panorama. As shown in Figure \ref{fig:alpha_choice}, our 360PanT with $\alpha=W$ demonstrates significantly better boundary continuity than the PnP baseline. However, upon closer inspection of the zoomed-in region indicated by the \textcolor{red}{red solid box}, a minor crack artifact is noticeable in the stitched area. Conversely, employing 360PanT with $\alpha=\frac{3W}{4}$ yields a 360-degree panorama without visible cracks in the stitched region. We set $\alpha$ to $\frac{3W}{4}$ in this paper. An intuitive explanation of this parameter choice, supported by further analysis, is provided in the supplementary material.

\subsection{Translation using Other Control Conditions}\label{subsection4.3}

To showcase the efficacy of our 360PanT (F) in handling diverse input conditions beyond 360-degree panoramic images, we present translated 360-degree panoramas generated from other control signals. Specifically, we consider a Canny edge map and a segmentation mask as the input control conditions, respectively, extracted from corresponding 360-degree panoramic images by using the same methods described in FreeControl \cite{freecontrol}. Figure \ref{fig:multiple} demonstrates a comparative study, highlighting the limitations of FreeControl in preserving visual continuity under these conditions. In contrast, our 360PanT (F) effectively maintains boundary continuity in the translated 360-degree panoramas.

\section{Conclusion}

We propose 360PanT, a training-free method for text-driven 360-degree panorama-to-panorama translation. This method integrates boundary continuity encoding and seamless tiling translation with spatial control. By constructing an extended input image, the boundary continuity encoding embeds continuity information from the original 360-degree panorama into a noisy latent representation. Guided by a target prompt, the seamless tiling translation with spatial control leverages this latent representation to generate a translated image with identical left and right halves while following the structure and semantic layout of the extended input image. This process successfully results in a final translated 360-degree panorama aligned with the target prompt. Extensive experiments on both real-world and synthesized 360-degree panoramas prove the effectiveness of our method in translating 360-degree panoramic images.


{\small
\bibliographystyle{ieee_fullname}
\bibliography{360pant}
}

\appendix

\section{Supplementary Content}

This supplementary material begins by providing an intuitive explanation for the choice of $\alpha$. Subsequently, we detail the process of producing target prompts for both real-world and synthesized datasets. Further visual results obtained under different control conditions are then presented. Finally, we showcase additional translated results from using different methods on real-world and synthesized 360-degree panoramic images.

\begin{figure*}[t]
\centering
\includegraphics[width=\textwidth]{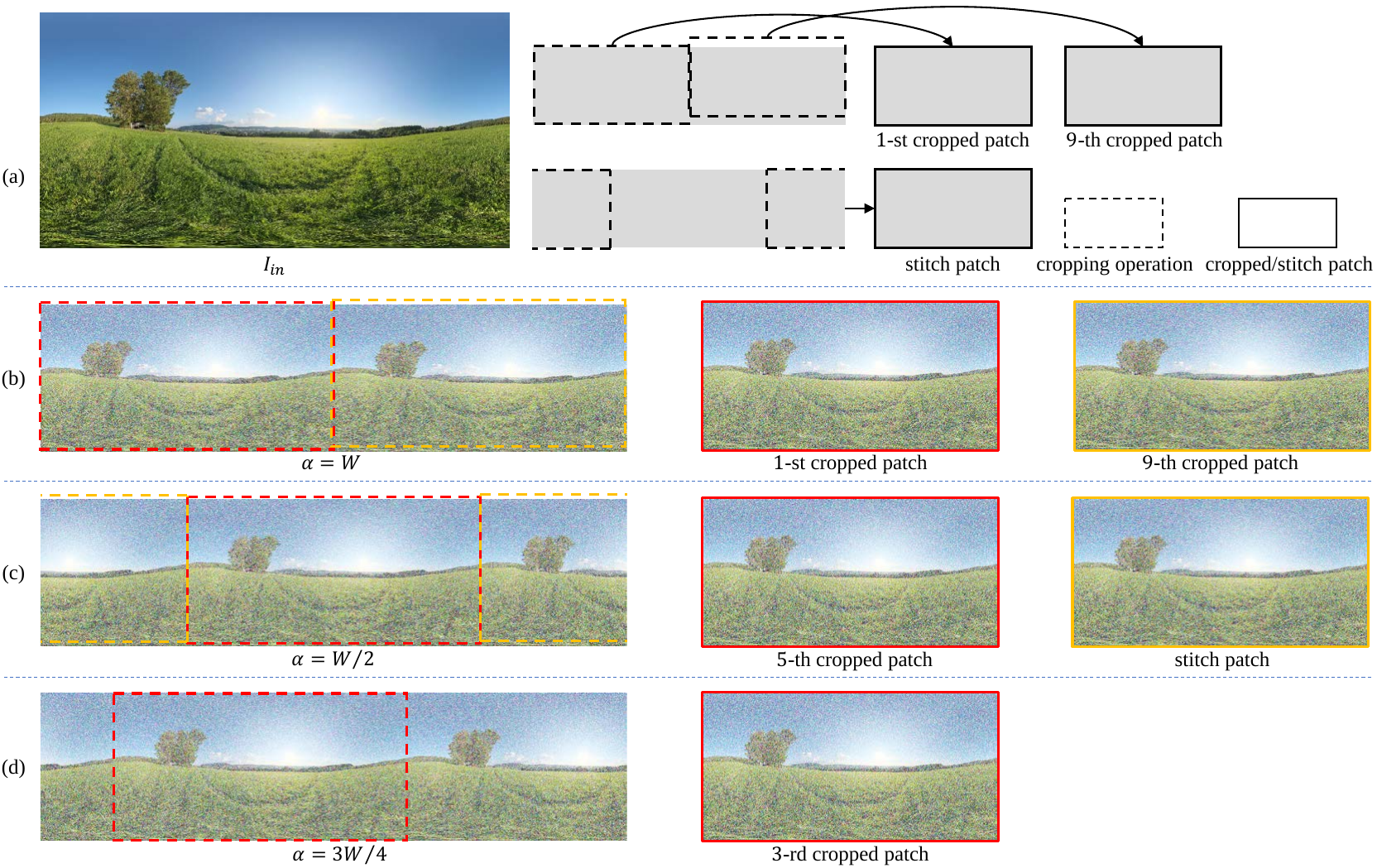}
\caption{\textbf{Intuitive explanation for the choice of split constant $\alpha$.} The cropped patches matching $I_{in}$ during the sliding window process are highlighted by \textcolor{red}{red} or \textcolor{yellow}{yellow} dashed boxes. Note that the stitch patch is a special cropped patch. At each denoising step $t$, when $\alpha=W$ in (b) or $\alpha=\frac{W}{2}$ in (c), two cropped patches matching $I_{in}$ but in different locations are denoised. Conversely, when $\alpha$ is set to $\frac{3W}{4}$ in (d), only one cropped patch matching $I_{in}$ undergoes denoising. To ensure better boundary continuity in the final translated result, we choose to set $\alpha$ to $\frac{3W}{4}$.}
\label{fig:alpha}
\end{figure*}

\begin{figure*}[t]
\centering
\vspace{5mm}
\includegraphics[width=\textwidth]{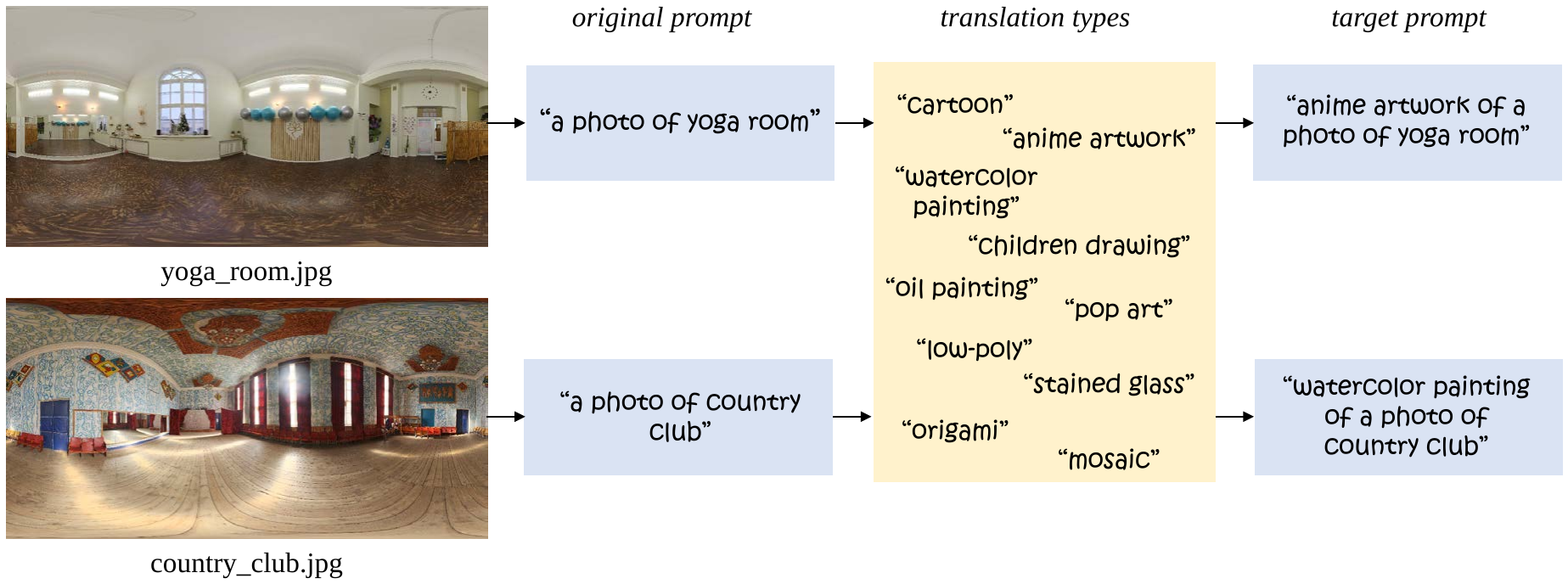}
\caption{\textbf{Target prompt generation for real-world 360-degree panoramas within the \emph{360PanoI-Pan2Pan} dataset.} Our 10 translation types are presented. A target prompt is formulated by combining a randomly selected translation type with the original prompt.}
\vspace{5mm}
\label{fig:data_g_real}
\end{figure*}

\noindent
\textbf{Explanation for the choice of $\alpha$.} To intuitively explain the choice of the split constant $\alpha$, Figure \ref{fig:alpha} visually depicts the cropping process in 360PanT at denoising step $t$ (where $t\in\{T, T-1, \cdots, 1\}$) for three distinct $\alpha$ values. The top row displays the input 360-degree panorama $I_{in}$ and a diagram of the cropping operations based on the sliding window mechanism employed in the seamless tiling translation with spatial control. Each cropped patch, including the special cropped patch (stitch patch), then undergoes independent denoising guided by a target prompt. Subsequent rows highlight the cropped patches matching $I_{in}$ during the sliding window process, indicated by \textcolor{red}{red} or \textcolor{yellow}{yellow} dashed boxes. Observe that when $\alpha=W$ or $\alpha=\frac{W}{2}$, two cropped patches matching $I_{in}$ but in different locations are denoised at each step $t$. Conversely, when $\alpha=\frac{3W}{4}$, only a single cropped patch matching $I_{in}$ undergoes denoising at each step.  Crucially, the continuity of boundaries of these highlighted patches are not considered during denoising. Consequently, at each denoising step $t$, the fewer cropped patches matching $I_{in}$ are denoised, the better the boundary continuity of the final translated 360-degree panorama. Therefore, we set $\alpha$ to $\frac{3W}{4}$ in this paper, which results in a final translated 360-degree panorama with seamlessly connected boundaries, effectively avoiding local visible cracks.

\begin{figure*}[t]
\centering
\vspace{-2mm}
\includegraphics[width=\textwidth]{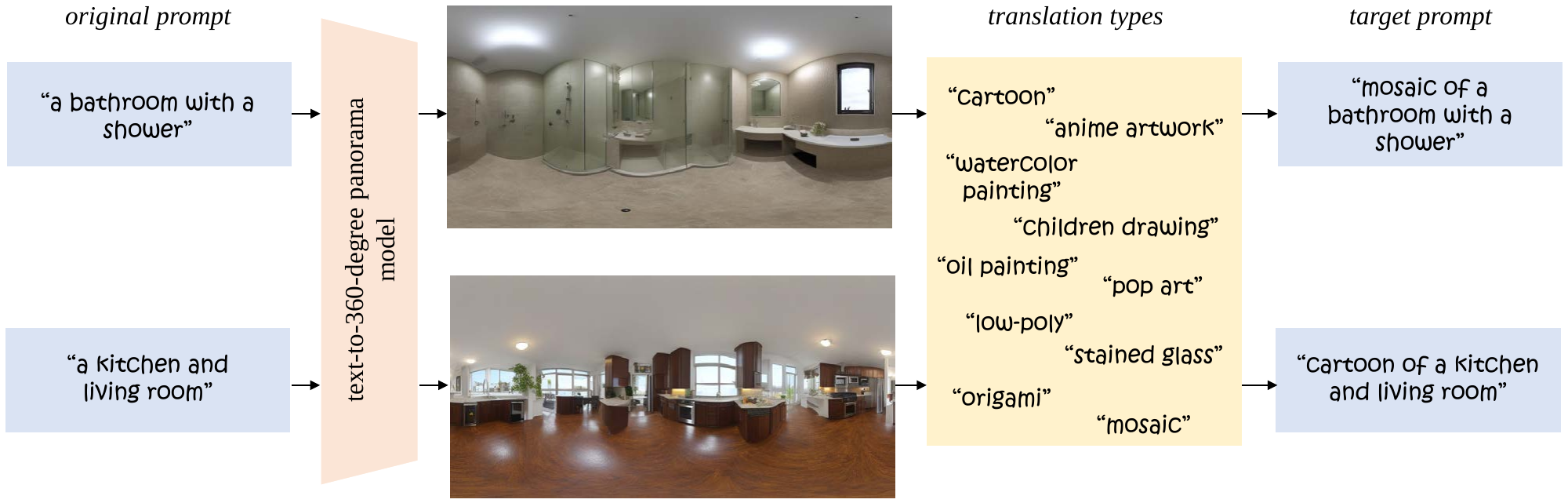}
\caption{\textbf{Target prompt generation for synthesized 360-degree panoramas in the \emph{360syn-Pan2Pan} dataset.} Each target
prompt consists of a randomly chosen translation type and
its corresponding original prompt.}
\label{fig:data_g_synthesized}
\end{figure*}

\noindent
\textbf{Generation process of target prompts.} Figure \ref{fig:data_g_real} illustrates the target prompt generation process for each real-world 360-degree panorama within the \emph{360PanoI-Pan2Pan} dataset. Utilizing a consistent template, ``a photo of $\{$image name$\}$", an original prompt is constructed for each 360-degree panoramic image. Subsequently, a target prompt is formulated by combining a randomly selected translation type with the original prompt. Figure \ref{fig:data_g_synthesized} depicts the analogous process for the \emph{360syn-Pan2Pan} dataset comprising synthesized 360-degree panoramas. Initially, 120 synthesized 360-degree panoramas are generated using a text-to-360-degree panorama model \cite{stitchdiffusion} guided by 120 original prompts. Similar to the real-world dataset, each target prompt consists of a randomly chosen translation type and its corresponding original prompt.

\begin{figure*}[t]
\centering
\vspace{-2mm}
\includegraphics[width=\textwidth]{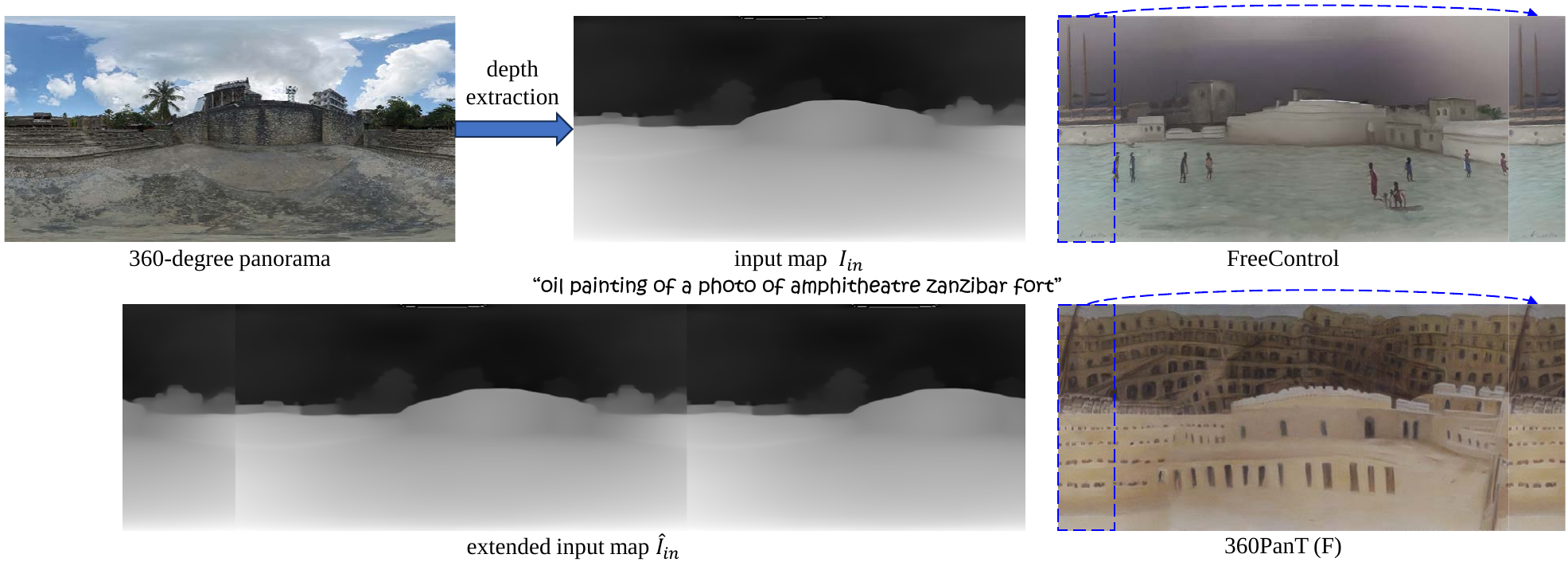}
\caption{\textbf{Depth map with discontinuous boundaries as the control condition.} The boundaries of depth map $I_{in}$ extracted from the 360-degree panorama are not continuous, resulting in visible cracks in the extended input map $\hat{I}_{in}$. In this situation, the translated panorama by our 360PanT (F) exhibits content inconsistency in the stitched area.}
\label{fig:depth}
\end{figure*}

\begin{figure*}[t]
\centering
\vspace{-2mm}
\includegraphics[width=\textwidth]{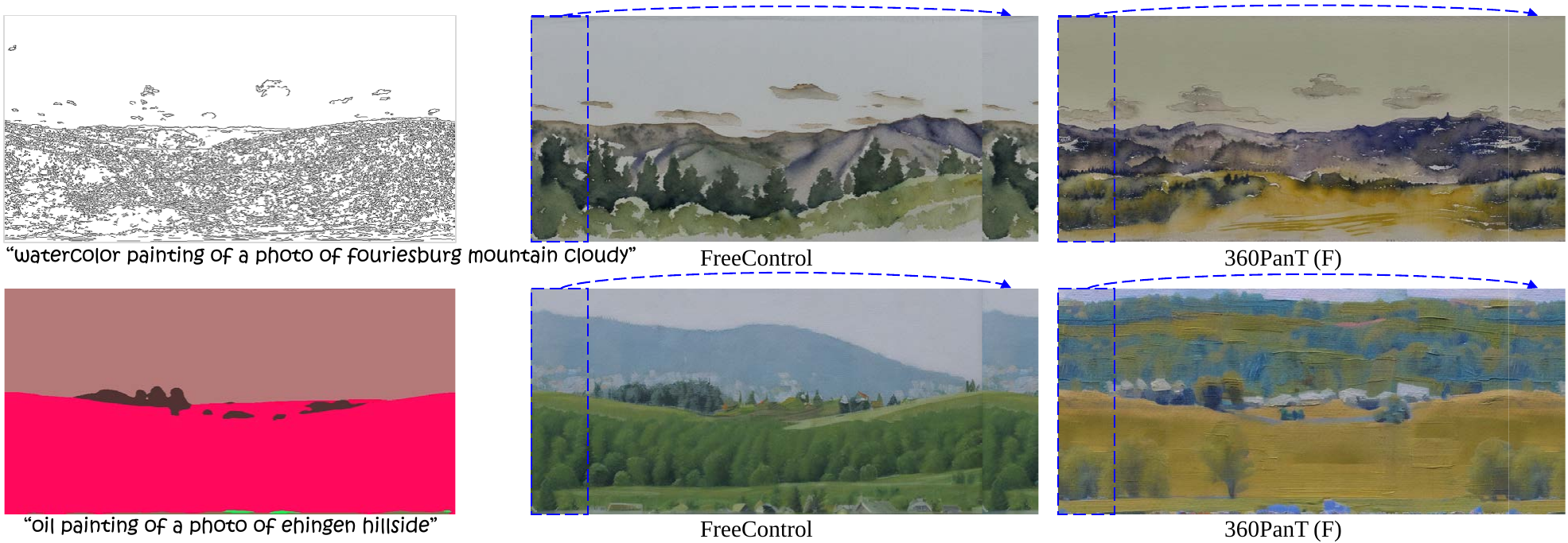}
\caption{\textbf{Visual results using other control conditions.} The extracted Canny edge map and segmentation mask can both effectively maintain continuous boundaries. When using them as control conditions, respectively, FreeControl is unable to guarantee the boundary continuity of the translated panoramas. In contrast, our 360PanT (F) enables the translated 360-degree panoramas with continuous boundaries regardless of the input conditions.}
\label{fig:multiple2}
\end{figure*}

\noindent
\textbf{Translation using other control conditions.} Diverse control conditions are extracted from corresponding 360-degree panoramic images using the methods described in FreeControl \cite{freecontrol}. If a control condition lacks continuous boundaries, the translated result by our 360PanT (F) will exhibit noticeable content inconsistency at the boundaries. For instance, Figure \ref{fig:depth} illustrates how using an extracted depth map $I_{in}$ with discontinuous boundaries as input leads to visible cracks in the extended input map $\hat{I}_{in}$. Consequently, the translated image by 360PanT (F) shows content inconsistency in the stitched area. In contrast, we observe that extracted Canny edge maps and segmentation masks effectively maintain continuous boundaries. As shown in Figure \ref{fig:multiple2}, when using them as control conditions, FreeControl fails to preserve boundary continuity, but our 360PanT (F) consistently produces translated 360-degree panoramas with continuous boundaries, regardless of the input conditions. 

\begin{figure*}[t]
\centering
\includegraphics[width=\textwidth]{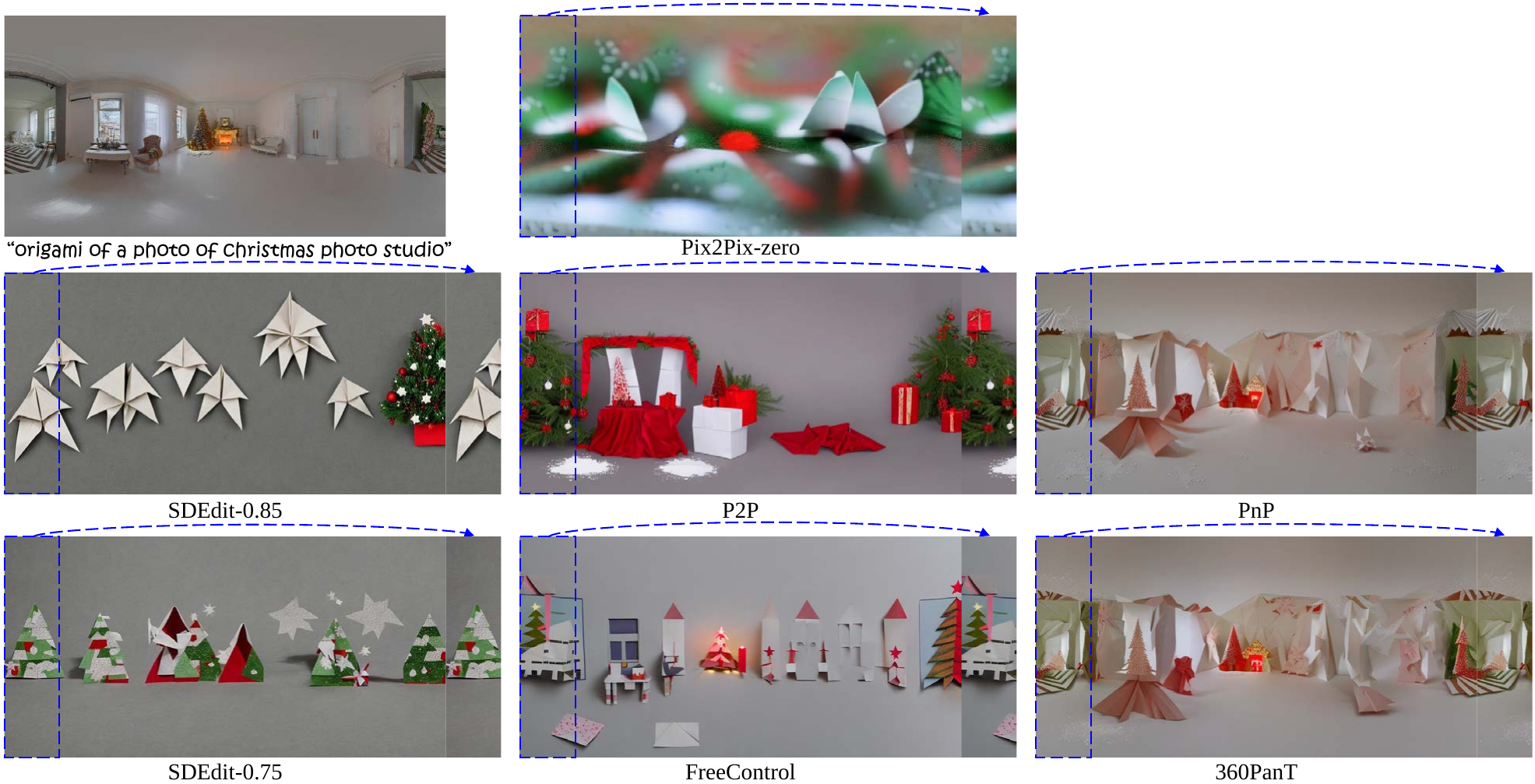}
\caption{\textbf{Visual results on real-world 360-degree panorama.} To easily identify visual continuity or discontinuity at the boundaries, we copy the left area of the panorama indicated by the \textcolor{blue}{blue dashed box} and paste it onto the rightmost side of the image.}
\label{fig:comp5}
\end{figure*}

\begin{figure*}[t]
\centering
\includegraphics[width=\textwidth]{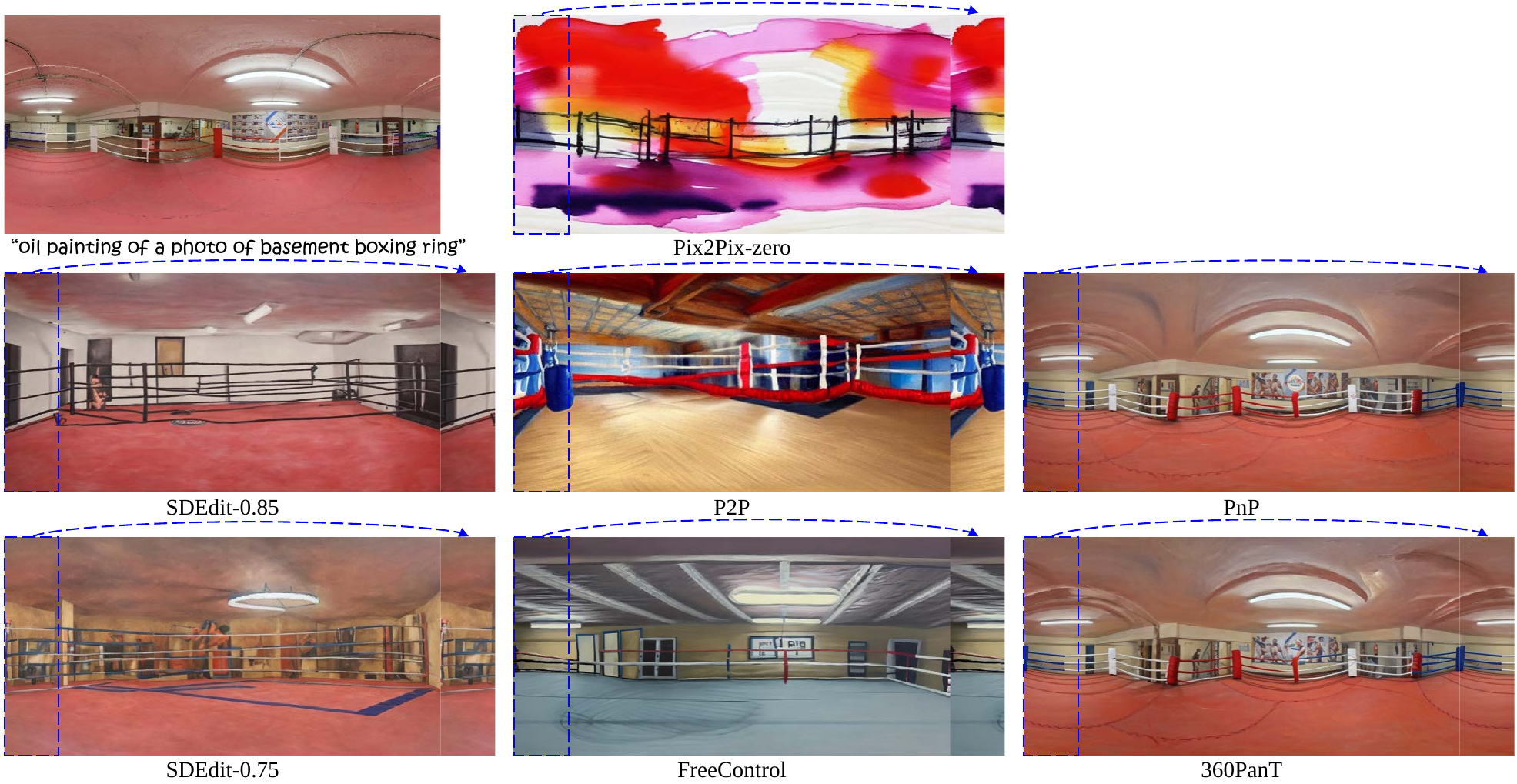}
\caption{\textbf{Visual results on real-world 360-degree panorama.}}
\label{fig:comp-real1}
\end{figure*}

\begin{figure*}[t]
\centering
\includegraphics[width=\textwidth]{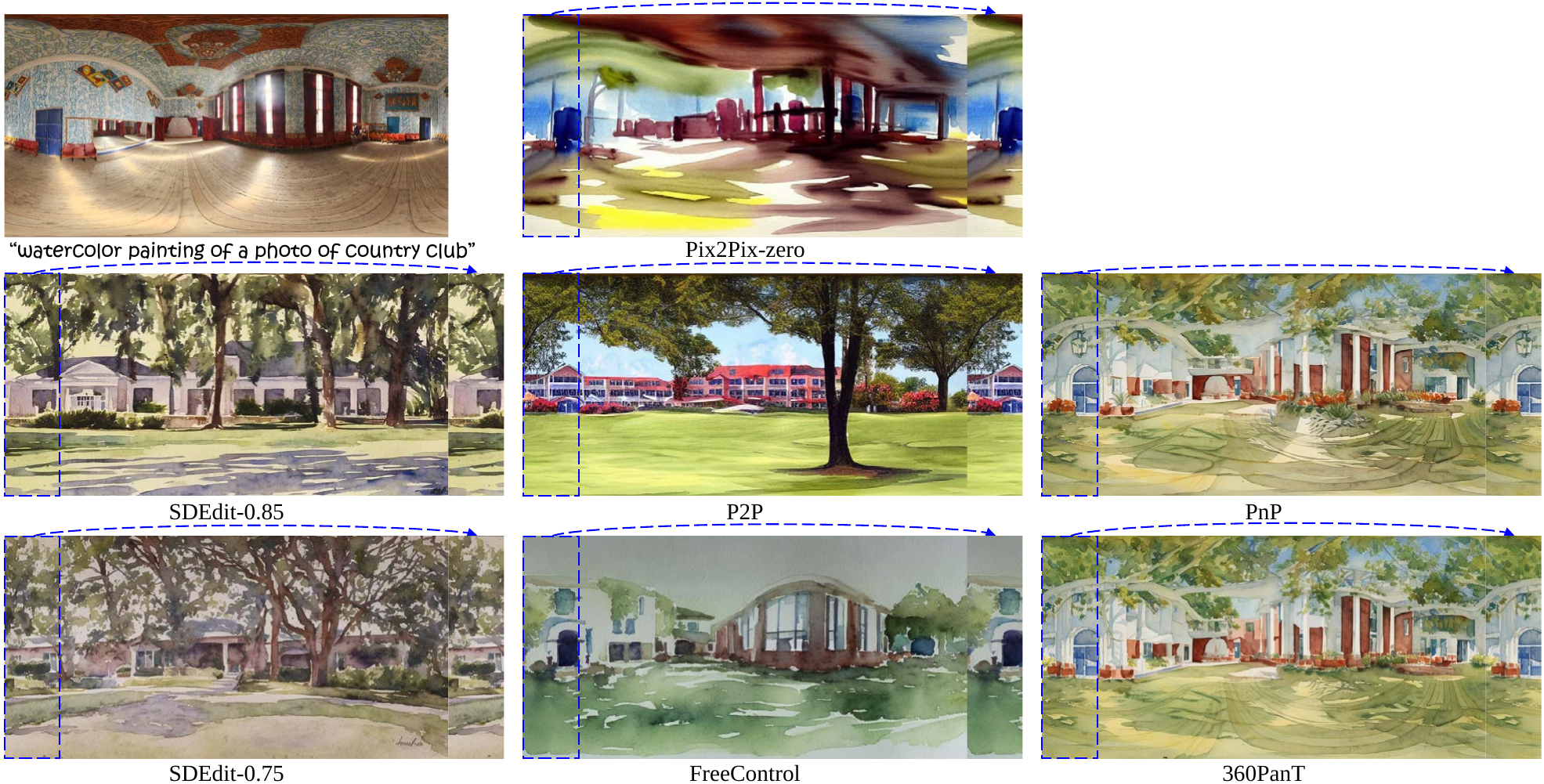}
\caption{\textbf{Visual results on real-world 360-degree panorama.}}
\label{fig:comp-real2}
\end{figure*}

\begin{figure*}[t]
\centering
\includegraphics[width=\textwidth]{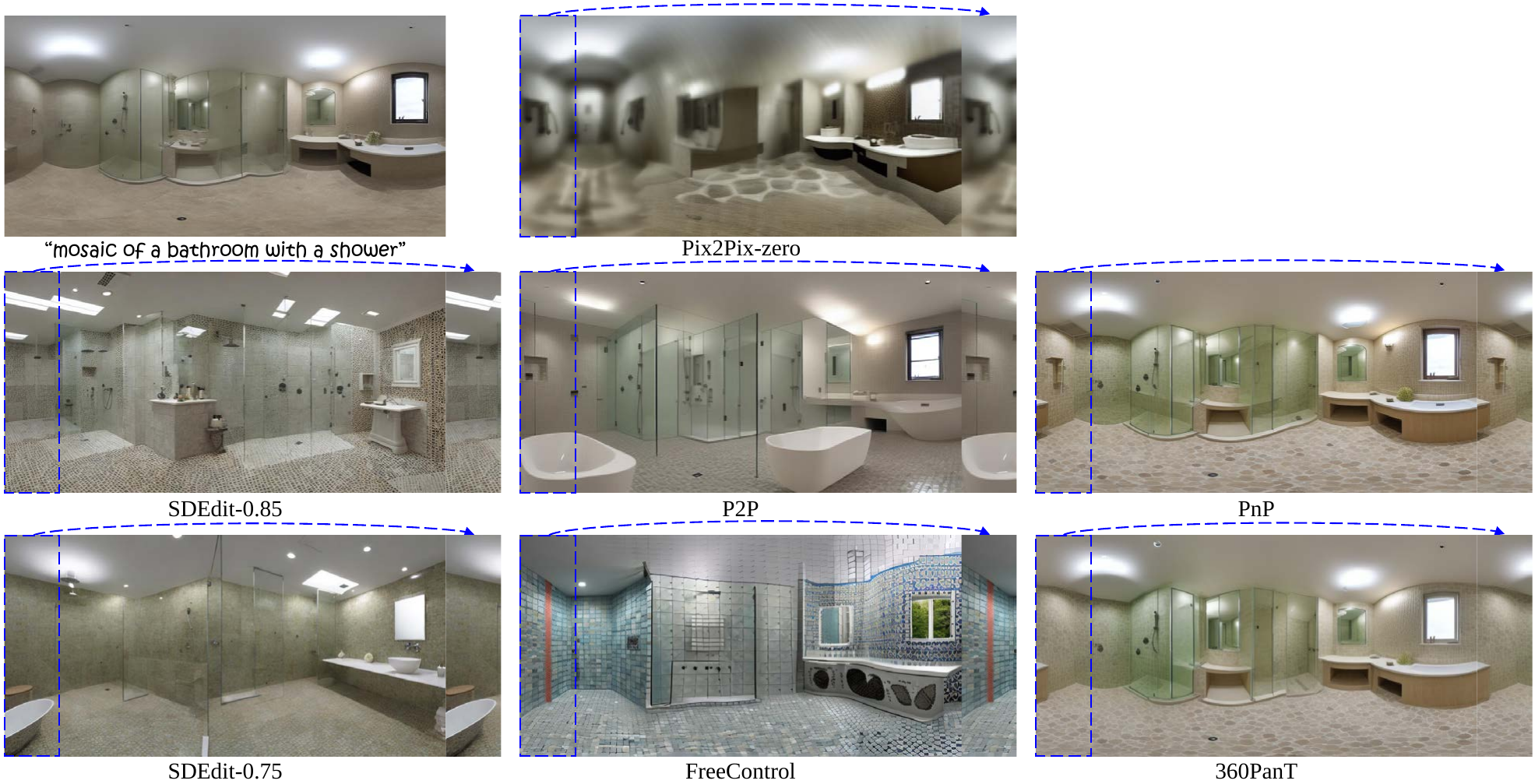}
\caption{\textbf{Visual results on synthesized 360-degree panorama.} To easily identify visual continuity or discontinuity at the boundaries, we copy the left area of the panorama indicated by the \textcolor{blue}{blue dashed box} and paste it onto the rightmost side of the image.}
\label{fig:comp8}
\end{figure*}

\begin{figure*}[t]
\centering
\includegraphics[width=\textwidth]{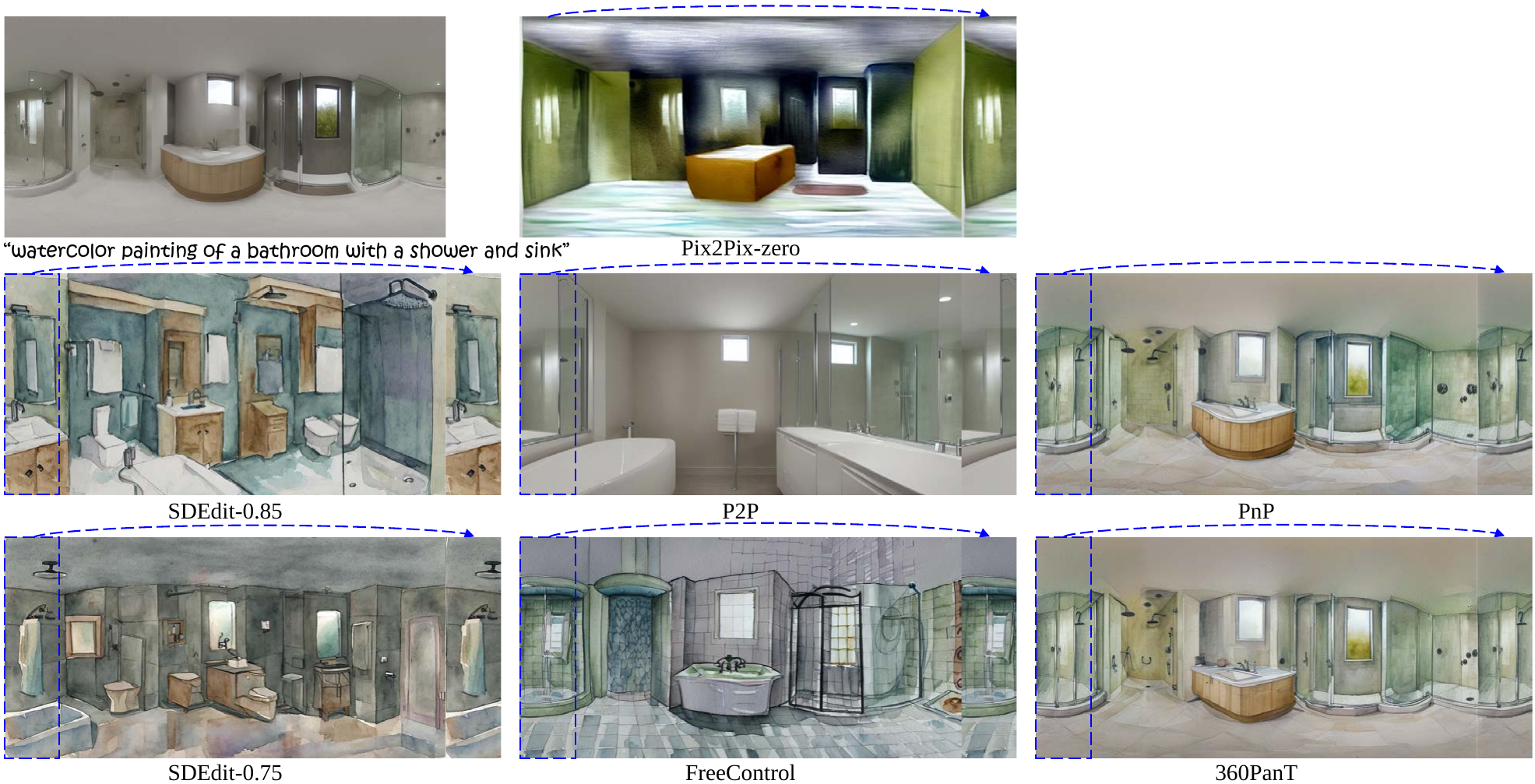}
\caption{\textbf{Visual results on synthesized 360-degree panorama.}}
\label{fig:comp6}
\end{figure*}

\begin{figure*}[t]
\centering
\includegraphics[width=\textwidth]{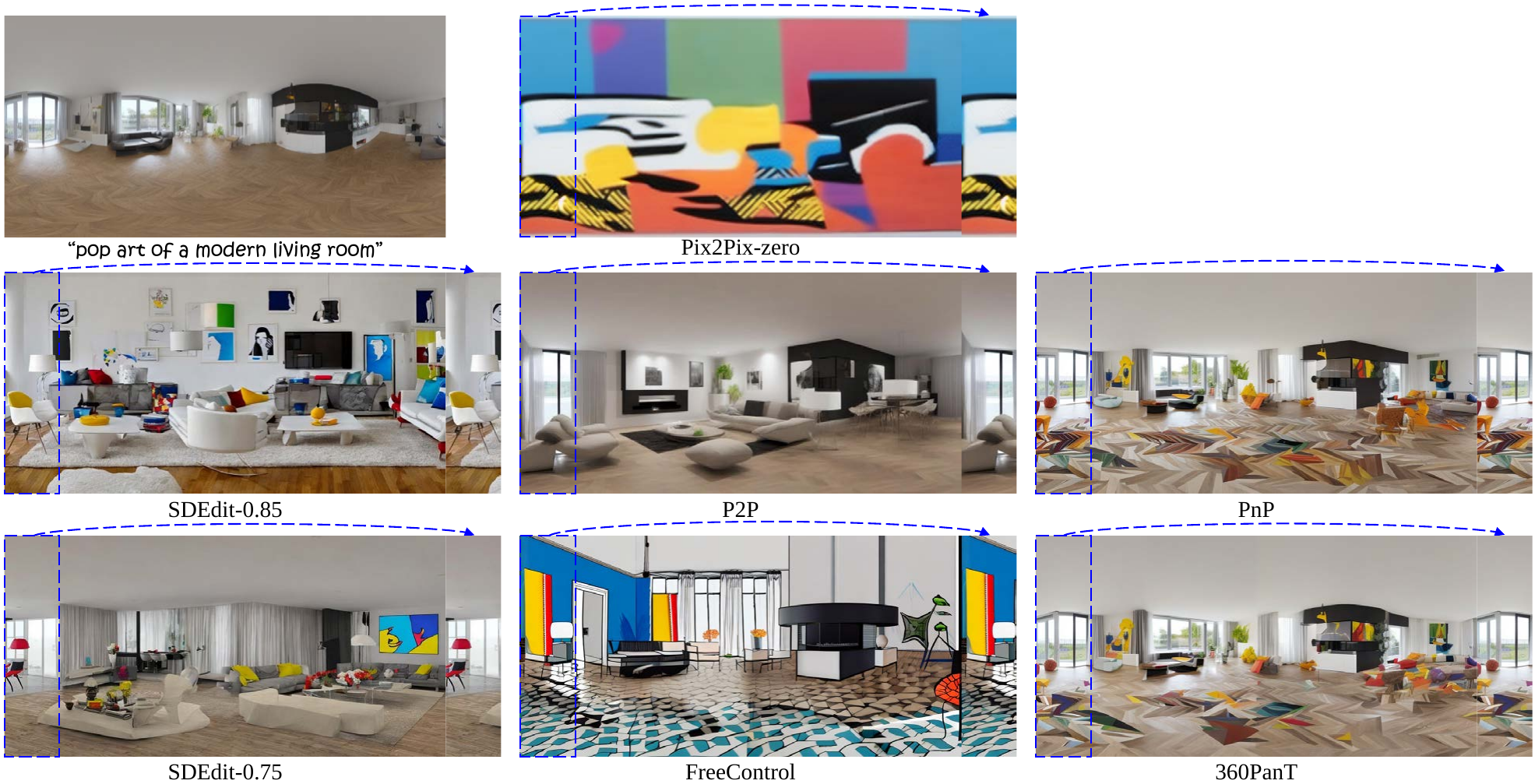}
\caption{\textbf{Visual results on synthesized 360-degree panorama.}}
\label{fig:comp18}
\end{figure*}

\noindent
\textbf{Visual results of various methods.} To further demonstrate the efficacy of 360PanT for 360-degree panorama translation, we present additional visual comparisons with SDEdit \cite{sdedit}, Pix2Pix-zero \cite{zeroi2i}, P2P \cite{P2P}, PnP \cite{PnP} and FreeControl \cite{freecontrol} on both real-world and synthesized 360-degree panoramas. As illustrated in Figures \ref{fig:comp5}, \ref{fig:comp-real1}, \ref{fig:comp-real2}, \ref{fig:comp8}, \ref{fig:comp6}, and \ref{fig:comp18}, 360PanT outperforms these methods in maintaining visual continuity at the boundaries while also adhering to the structure and semantic layout of the input 360-degree panoramic images.

\end{document}